\documentclass{article} 
\usepackage{geospatialbench_preprint,times}
\usepackage{iftex}
\ifXeTeX

  \usepackage{fontspec}
\fi

\usepackage{amsmath,amsfonts,bm}

\def\eqref#1{equation~\ref{#1}}

\def\1{\bm{1}}

\DeclareMathAlphabet{\mathsfit}{\encodingdefault}{\sfdefault}{m}{sl}
\SetMathAlphabet{\mathsfit}{bold}{\encodingdefault}{\sfdefault}{bx}{n}

\newcommand{\ArxivPdfAuthors}{Zheng Ren, Yanning Wang, Ziheng Huang, Fei Huang, Shiqi Wang, Ziqian Ye, Yang Yue}

\author{%
\makebox[\dimexpr\textwidth-2\tabcolsep\relax][c]{%
\begin{minipage}{0.94\textwidth}
\centering
\begin{tabular*}{\linewidth}{@{\extracolsep{\fill}}cccc@{}}
\textbf{Zheng Ren\textsuperscript{\textdagger}} &
\textbf{Yanning Wang\textsuperscript{\textdagger}} &
\textbf{Ziheng Huang\textsuperscript{\textdagger}} &
\textbf{Fei Huang}
\end{tabular*}\\[0.25em]
\textbf{Shiqi Wang \quad Ziqian Ye \quad Yang Yue\textsuperscript{*}}\\[0.45em]
\normalfont Urban Governance and Design (UGOD) Thrust\\
\normalfont The Hong Kong University of Science and Technology (Guangzhou)\\
\normalfont Guangzhou, China\\
\normalfont\texttt{yangyue@hkust-gz.edu.cn}\\[0.15em]
\normalfont\small \textsuperscript{\textdagger} Equal contribution. \textsuperscript{*} Corresponding author.
\end{minipage}%
}%
}

\usepackage{hyperref}
\hypersetup{
  hidelinks,
  hypertexnames=false,
  pdftitle={Mind the Spatial Gap: A Benchmark for Geospatial Reasoning and Constrained Spatial Editing},
  pdfauthor={\ArxivPdfAuthors},
  pdfsubject={Preprint},
  pdfkeywords={geospatial reasoning, spatial relations, multimodal large language models, spatial editing, benchmark}
}
\usepackage{url}
\usepackage{graphicx}
\usepackage{booktabs,float,placeins}

\title{Mind the Spatial Gap: A Benchmark for Geospatial Reasoning and Constrained Spatial Editing}

\begin{document}

\maketitle

\begin{abstract}
Understanding and manipulating spatial relations is fundamental to AI systems that reason about physical environments and construct grounded world models. However, existing evaluations of multimodal large language models (MLLMs) for spatial reasoning rely predominantly on static question-answering (QA) over small-scale or synthetic scenes, rarely probing whether models can modify spatial configurations while satisfying explicit geometric constraints. We introduce GeoSpatialBench, a benchmark designed to evaluate an escalating progression of spatial cognition, ranging from spatial relation understanding and geometric reasoning to constrained spatial editing. Grounded in real-world urban data across six geographically diverse cities, GeoSpatialBench evaluates a comprehensive taxonomy of 28 topological, directional, metric, and ordering relations across structured GeoJSON and map image modalities, with ground-truth references and constraints verified via GIS operations. Our empirical evaluation across frontier MLLMs uncovers a stark spatial gap: structured GeoJSON consistently outperforms map images, with multimodal combinations offering no reliable gain over GeoJSON alone, and while models show reasonable competence on qualitative topological and ordering, performance drops on quantitative metric relations. Crucially, editing performance on ordering relations declines relative to QA, and most failed edits violate at least one specified spatial constraint. These results highlight a gap between passively recognizing spatial relations and actively manipulating geometric configurations. GeoSpatialBench provides an open, rigorous diagnostic testbed to benchmark these spatial limitations, laying essential empirical groundwork for physically grounded world models and embodied spatial AI.

\end{abstract}

\section{Introduction}
\label{sec:geospatialbench-introduction}

The ability to understand spatial relations and use them to manipulate spatial configurations is fundamental to AI systems that reason about physical environments and construct world models \citep{ha2018world,assran2025vjepa2}. Consider planning a road across a region: it needs to avoid mountains and rivers, cross a densely populated district, remain connected, and meet a distance limit. Such reasoning underlies navigation, autonomous driving, spatial planning, embodied interaction, and world models \citep{Zhou2025DINOWM,Hong2026ESIBench}. Yet existing work on world models, embodied AI, and multimodal spatial reasoning is primarily evaluated in confined indoor or local scenes and over short trajectories \citep{Cheng2024SpatialRGPT,Yang2025ThinkingInSpace,Hong2026ESIBench}. Geospatial reasoning at city and regional scales introduces a different challenge: models must reason over heterogeneous geographic features and their topological, directional, metric, and ordering relations across larger spatial extents and multiple scales.

Recent multimodal large language models (MLLMs) have achieved impressive results on visual/geometric benchmarks. The proprietary GPT-6 Astra scores 95.9\% with tools on BenchCAD, which evaluates programmatic CAD reasoning \citep{zhang2026benchcad,OpenAI2026GPT6Astra}. The open-weight Kimi K3 scores 81.6\% without tools on MMMU-Pro, a benchmark of multidisciplinary visual understanding and reasoning \citep{yue-etal-2025-mmmu,KimiTeam2026KimiK3}. These results demonstrate strong general spatial and visual reasoning capabilities but provide limited evidence about geospatial reasoning over real-world geographic configurations. Recent studies have begun to explore this problem using different forms of geospatial representation, including maps \citep{dihan2025mapeval,ung2025cartomapqa}, imagery \citep{xu2026earthspatialbench}, vector geometry \citep{bao2026urbangeoeval}, and structured urban data \citep{ji2025foundation}. Across these settings, evaluations cover spatial relation families such as topological, directional, metric and ordering. However, the extent to which models can consistently recognize these relations and use them to modify spatial configurations remains unclear.

Most existing benchmarks use question-answering (QA) to assess whether models can understand and apply the underlying spatial relations. A correct answer, however, may not show whether that relation can guide a change to the scene \citep{vafa2024worldmodel}. More broadly, benchmark answers need not reflect coherent use of the underlying concepts \citep{mancoridis2025potemkin}. Constrained spatial editing provides a complementary solution: object placements, geometric constructions, and urban layouts can be directly verified against explicit constraints \citep{abdelreheem2025placeit3d,huang2025fireplace,kim2026geobuildbench,liu2026cityediting}. Yet, existing studies usually evaluate editing separately from QA, rather than testing whether models can both recognize and apply the same spatial relations.

We introduce GeoSpatialBench, a benchmark that combines QA tasks on spatial relations with constrained spatial editing (Figure~\ref{fig:geofm-pipeline}). By evaluating both tasks against the same set of spatial relations, GeoSpatialBench provides a stronger test of whether models can apply the relations they identify. It covers 28 formally defined relations in four families (topological, directional, metric, and ordering), drawing on established Geographic Information Science (GIS) and spatial-database concepts \citep{egenhofer1991categorizing,ogc2011sfa}. The benchmark contains 8,400 instances derived from real OpenStreetMap data across six cities. QA and editing use separate instance pools but share the same set of relations. Each scene is available as GeoJSON or as a map rendered from the same geometry. Our results show that GeoJSON generally supports stronger geospatial reasoning, whereas map images remain substantially weaker. This gap highlights an underexplored challenge in using maps as a visual medium for precise geospatial reasoning.

Our contributions are threefold. First, we introduce GeoSpatialBench, a comprehensive geospatial relation benchmark covering 28 spatial relations recognized in spatial database and GIS, with instances drawn from six cities worldwide. Second, we propose spatial editing with QA tasks to verify models' genuine spatial reasoning abilities. Third, we systematically compare leading models, including GPT-6 Astra, across input modalities using GIS-derived answers and deterministic checks of edited geometry. Together, these contributions enable more accurate measurement of spatial-relation understanding and reveal where current models still need improvement. Section~\ref{sec:related-work} reviews related work, Section~\ref{sec:methods} describes the benchmark and evaluation protocol, Section~\ref{sec:experiments} presents the results and analyses, and Section~\ref{sec:conclusion} concludes.

\section{Related Work}
\label{sec:related-work}
\subsection{Spatial relations and spatial reasoning of AI systems}

Spatial relations describe how points, lines, and regions are arranged and relate to one another. GIS and qualitative spatial reasoning formalize a broad range of such relations and provide geometric predicates for evaluating them \citep{egenhofer1991pointset,frank1992qualitative,cohn2008qualitative,ogc2011sfa}. Recent work has extended spatial reasoning to MLLMs. SpatialRGPT uses region and depth information to reason about relative direction and distance, SpatialReasoner incorporates explicit 3D representations, and SpatialScore evaluates a wider range of visual spatial abilities \citep{Cheng2024SpatialRGPT,Ma2025SpatialReasoner,Wu2026SpatialScore}. Related work on world models uses learned representations for prediction and planning, while VSI-Bench and ESI-Bench examine spatial recall and evidence gathering through action \citep{ha2018world,assran2025vjepa2,Zhou2025DINOWM,Yang2025ThinkingInSpace,Hong2026ESIBench}. Together, these studies establish the importance of spatial knowledge across visual and embodied settings, but their evaluations largely center on local scenes with limited spatial extent. It remains less explored whether current MLLMs can reason over a broad set of formal spatial relations among georeferenced geographic features, derive geometric quantities from their configurations, and use those relations to modify spatial configurations.

\subsection{Benchmarks for geospatial reasoning in MLLMs}

Recent benchmarks have extended spatial reasoning to larger-scale geographic settings and real-world geospatial data. MapEval and CartoMapQA assess map reading and route reasoning, while FRIEDA evaluates multi-step reasoning over map images \citep{dihan2025mapeval,ung2025cartomapqa,pyo2026frieda}. EarthSpatialBench combines Earth imagery with explicit geographic coordinates to evaluate distance, direction, and topology, while UrbanGeoEval provides raw city-scale geometry for spatial inference \citep{xu2026earthspatialbench,bao2026urbangeoeval}. \citet{ji2025foundation} further study topological inference from WKT and relation-conditioned geometry generation. These studies demonstrate that geospatial reasoning can be evaluated using real geographic data and a range of representations, including map images, imagery, and structured or vector geometry. However, existing benchmarks differ substantially in their representations, relation sets, and task formulations, making it difficult to compare how models reason across spatial modalities and relation families. In particular, few systematically compare GeoJSON with a map image rendered from the same underlying geometry while evaluating a broad and formally defined set of spatial relations.

\subsection{Spatial editing and relation-grounded evaluation}

Most existing geospatial benchmarks evaluate models via QA tasks, testing their ability to recognize spatial relationships and compute geometric quantities. However, obtaining correct answers does not prove that a model can leverage these relationships to modify spatial configurations. Benchmark success can mask gaps in a model's underlying understanding \citep{vafa2024worldmodel,mancoridis2025potemkin}. Spatial editing provides a complementary test because the resulting configuration can be checked against explicit spatial constraints. Recent work has explored language-guided object placement in 3D scenes, executable CAD editing, and iterative modification of urban GeoJSON \citep{abdelreheem2025placeit3d,zhang2026benchcad,liu2026cityediting}. These studies show the feasibility of evaluating whether models can translate spatial instructions into valid geometric changes. However, spatial editing has generally been studied separately from geospatial QA, rather than evaluating whether models can recognize and then apply the same spatial relations.

GeoSpatialBench brings the two together, scoring each task by relation family on separate instance pools and verifying answers and edits with GIS operations. This design provides a direct test of whether such spatial reasoning ability transfers to editing geographic geometry so that the models actually understand the underlying spatial relations.

\section{GeoSpatialBench}
\label{sec:methods}
\begin{figure}[t]
  \centering
  \includegraphics[width=\linewidth]{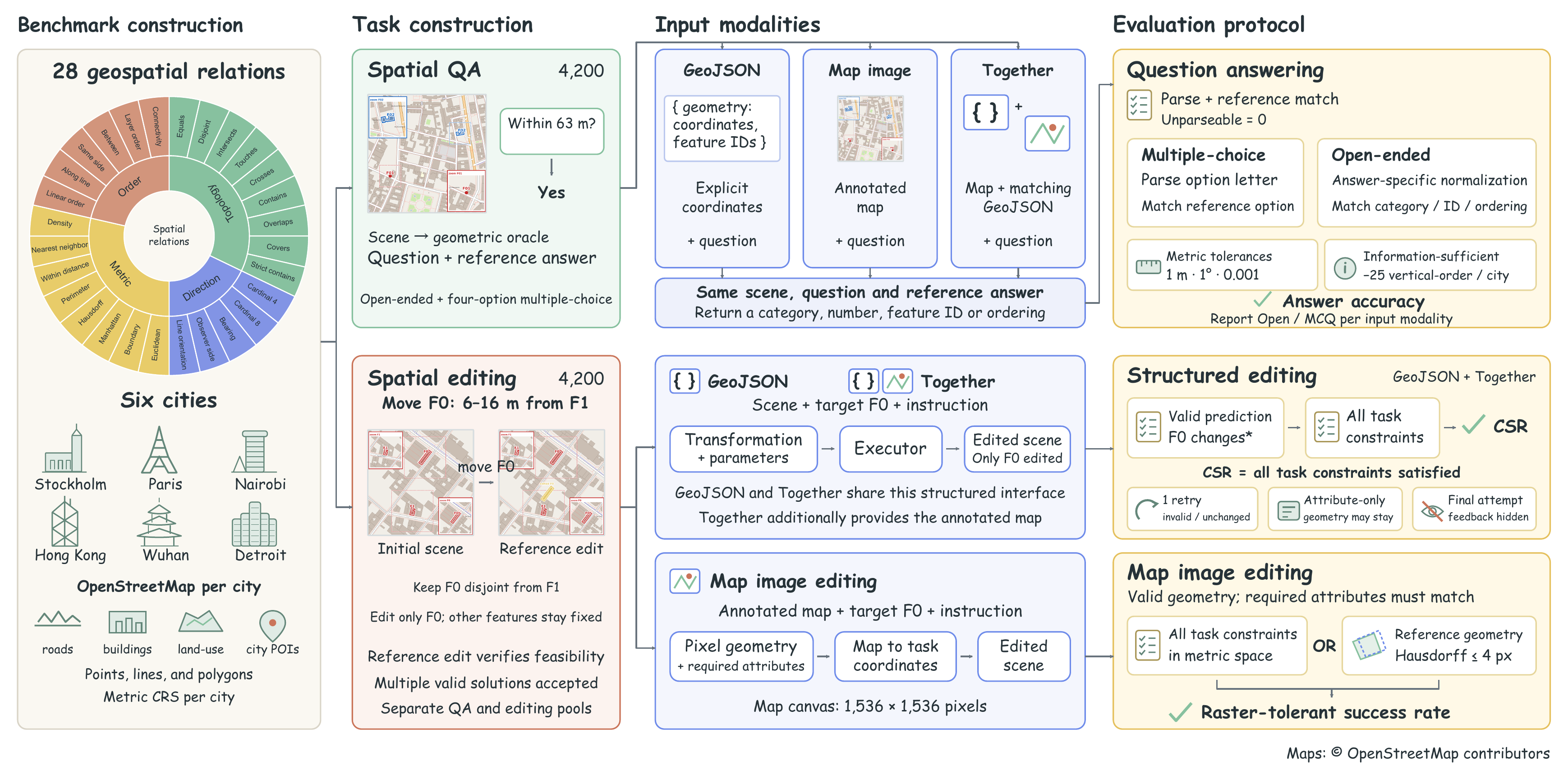}
  \caption{Overview of GeoSpatialBench. QA and spatial editing are constructed from
urban vector data across six cities and 28 spatial relations. Both tasks support
GeoJSON, image, and combined inputs. QA is evaluated by answer
accuracy. GeoJSON and Together editing use the Constraint Satisfaction Rate
(CSR), while map-image editing uses a raster-tolerant success rate.}
  \label{fig:geofm-pipeline}
\end{figure}

GeoSpatialBench evaluates spatial reasoning through spatial question answering and
constrained geometric editing. Figure~\ref{fig:geofm-pipeline} summarizes the benchmark construction, model inputs, and evaluation procedures.

\subsection{Task Definition}

GeoSpatialBench covers 28 geospatial relations over points, lines, and polygons at the city scale. These relations are categorized into four groups: topology, direction, metric, and order. Instances of these relations include touching, relative direction, distance, and connectivity. Each relation is defined by a precise geometric or attribute-based rule. For instance, two roads are considered connected only if they share an endpoint, while vertical order is ascertained by road-layer attributes.

In QA, models are tasked with providing an answer based on a specific scene and an associated question. Depending on the types of relations involved, the answer may take the form of a category, a number, a feature ID, or an ordering. Each QA instance is presented in both open-ended and four-option multiple-choice formats.

Spatial editing requires the model to modify the scene instead of describing it. Given an initial scene, an instruction, and a target feature $F_0$, the model is expected to make edits exclusively to the specified feature $F_0$, while leaving other features unchanged. Editing accuracy is assessed by whether the scene resulting from the model's edit satisfies the requested spatial constraints. The aim of the editing task is to evaluate whether the models truly comprehend spatial relations.

\subsection{Benchmark Construction}

The benchmark covers six cities with diverse geographic contexts: Stockholm,
Paris, Hong Kong, Wuhan, Detroit, and Nairobi. Each city contributes 700 QA and 700 editing instances, giving 8,400 canonical instances in total.

We use OpenStreetMap data for roads, buildings, land-use regions, and
city-level points of interest (POIs). All features are projected to a
city-specific metric coordinate system before distance and area calculations.
We then filter them by geometry type and city-specific length or area
thresholds. Road-layer attributes are retained for vertical-order questions.
QA and editing use separate feature pools, so their instances are not paired.

For each geospatial relation, we construct a canonical scene from the filtered
features. The scene records the required geometries, roles, source attributes,
and construction provenance. It may use observed geometries or controlled
constructions when the source data do not provide enough cases. A geometric
oracle evaluates the scene and produces the reference answer. Source identities
and construction records are kept for auditing but excluded from model inputs.

Each canonical scene is serialized as a GeoJSON FeatureCollection and also
rendered as an annotated map. GeoJSON preserves the task-coordinate geometry
and uses anonymous feature identifiers, with only task-required properties.
For the map, the same scene is converted to WGS84 and rendered in a square Web
Mercator view. Task features are highlighted and labeled, with a legend,
detail insets, and attribute cards added when needed. All images use a
$1{,}536 \times 1{,}536$ pixel canvas. 

QA questions are generated from the oracle output. Open-ended questions ask
the model to return the answer in the required form. Multiple-choice
distractors depend on the relation. Among all, directional relations use nearby bearings, ordering relations use alternative
sequences, and numerical relations use adjacent intervals. Topological
relations and feature-selection tasks use plausible competing labels or
feature IDs. Each item is checked before use. We verify the geometry and reference answer, then check that the options are distinct.

Editing instances are constructed separately from QA. Candidates are rejected
if the spatial constraints specified by the instruction are already satisfied
in the initial scene, and retained only when a reference edit confirms that a
valid solution exists. The reference edit is used only to verify feasibility;
any valid edit that satisfies the instruction is accepted. Geometric edits
must change the target feature $F_0$, while attribute edits may leave its geometry
unchanged.

\subsection{Input Modalities}

We evaluate three input modalities: GeoJSON, map image, and their combination
(Together). GeoJSON provides explicit coordinates and anonymous feature
identifiers, while map image shows the same scene as an annotated map. Together
pairs the annotated map with the same GeoJSON used in the GeoJSON condition.
For QA, all three conditions use the same scene, question, and reference answer,
with both open-ended and four-option multiple-choice forms. The answer format
remains unchanged across conditions, allowing us to compare the effect of input
representation on spatial reasoning.

For editing, GeoJSON uses a structured interface. Models receive the scene, target feature $F_0$, and instruction, and return a transformation and its parameters (e.g., translation, rotation, scaling, or attribute editing). An executor applies the transformation to $F_0$, and the evaluator checks the resulting scene against the required spatial constraints. Map image editing instead returns edited geometry in pixel coordinates with any required attributes. The evaluator maps the prediction back to the task coordinate system before checking the same constraints. All images use a $1{,}536 \times 1{,}536$ pixel canvas.

Together follows the same structured editing interface as GeoJSON while additionally providing the corresponding annotated map. This allows a direct comparison with GeoJSON to test whether visual information improves structured editing. By contrast, map image editing changes both the input representation and the output interface because the model must generate pixel-coordinate geometry. Differences between GeoJSON and map image editing should therefore not be interpreted as a pure input-only ablation.

\subsection{Evaluation Protocol}

\paragraph{Question answering.}
We use deterministic scoring rather than a learned judge. Multiple-choice outputs
are parsed as option letters, while open-ended outputs use answer-specific
normalization. Distance and perimeter answers allow a one-meter tolerance,
bearing answers a one-degree circular tolerance, and density answers a tolerance
of 0.001 in the stated units. Unparseable answers are counted as incorrect, and
open-ended and multiple-choice accuracy are reported separately.

GeoJSON, map image, and Together use the same canonical scene, question, and
reference answer. GeoJSON exposes the exact task-coordinate geometry, map image
shows its annotated rendering, and Together provides both. For geometry-only
conditions, we also report an information-sufficient score that removes the 25
vertical-order questions per city whose required layer attributes are not
provided. Annotated maps may provide these attributes through attribute cards,
but raster measurements still limit visual precision.

\paragraph{Structured editing.}
GeoJSON and Together share the same structured editing protocol. Given an
initial scene, target feature $F_0$, and instruction, the model returns a
transformation and its parameters. Let $S_i$ and $S'_i$ denote the initial and
edited scenes, and let $\mathcal{C}_i$ be the task constraints. Let $V_i$
indicate a parseable, nonempty, valid prediction. Let $U_i$ indicate that a
geometric edit changes $F_0$; for attribute-editing tasks, we set $U_i=1$.
We define
\begin{equation}
E_i =
V_i U_i
\prod_{c\in\mathcal{C}_i}
\mathbf{1}\!\left[c(S_i,S'_i)\right],
\qquad
\mathrm{CSR} =
\frac{1}{N}\sum_{i=1}^{N} E_i .
\end{equation}
Here, $E_i$ is a binary indicator of whether edit $i$ is valid and satisfies
all task constraints, and Constraint Satisfaction Rate
(CSR) is the proportion of successful edits over the
evaluation set. With one corrective retry, an undecodable or unchanged edit triggers a second attempt requesting a valid, nonzero edit. We score only the final attempt; the
feedback does not reveal whether the task constraints were satisfied.

\paragraph{Map image editing.}
The evaluator first maps the predicted pixel geometry back to the task
coordinate system. It then accepts either metric-space constraint satisfaction
or pixel geometry within four pixels of the reference under Hausdorff distance,
provided that required attributes match. We report this raster-tolerant success
rate separately from strict CSR. The reference geometry is used for feasibility
verification and image-level scoring.

We aggregate scores by task, input modality, and QA form, and report results by
city, relation family, and individual relation. Comparisons across GeoJSON, map
image, and Together use matched scenes and instructions whenever the model
supports the corresponding input modality.

\section{Evaluation on GeoSpatialBench}
\label{sec:experiments}

\subsection{Evaluation Setup}
\label{sec:evaluation-setup}

We evaluate nine models across the six cities, including four open-weight
models (Qwen3.5-9B, Qwen3.8-27B, Qwen3-VL-8B-Instruct, and InternVL3-8B)
and five API-accessed models (GPT-6 Astra, GPT-5.6 Sol, KIMI K3,
MiMo-V2.6-Pro, and DeepSeek-V4.1-Flash).

For question answering, each model is evaluated in both open-ended and
four-option multiple-choice settings. We consider three input settings:
GeoJSON, map image, and Together, where Together combines the GeoJSON
representation with the corresponding annotated map. Spatial editing is
evaluated under the same three input settings using the task-specific output
interfaces described in Section~\ref{sec:methods}. We also report performance separately for open-ended QA, multiple-choice QA, and spatial editing. Additional model- and setting-specific results are provided in Appendix~\ref{app:additional-results}.

\subsection{Overall Performance}

\begin{figure}[t]
    \centering
    \includegraphics[width=\linewidth]{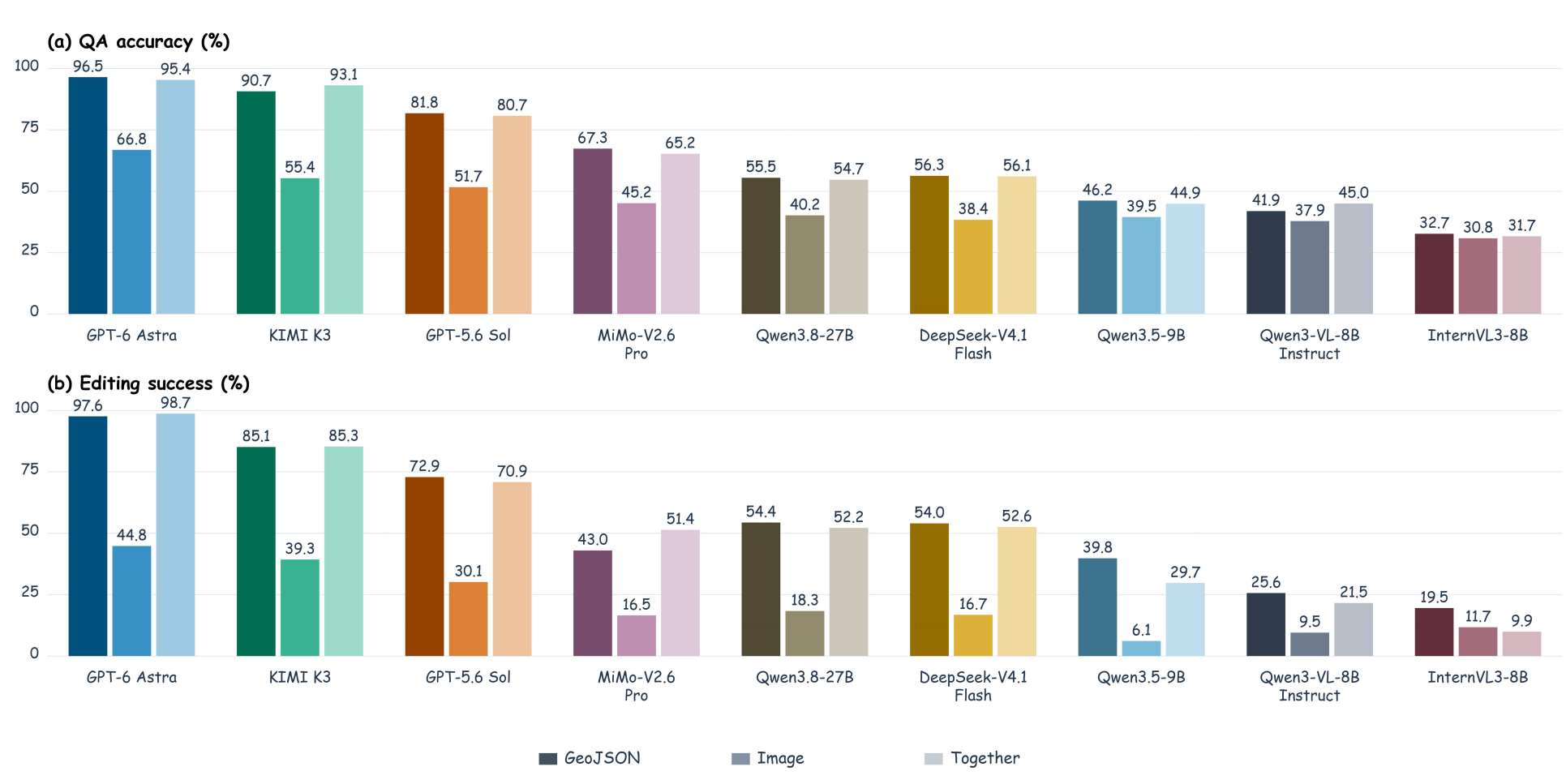}
    \caption{
    Average QA and spatial editing performance across models and input modalities.
    (a) QA accuracy averaged over open-ended and multiple-choice settings.
    (b) Spatial editing success. GeoJSON and Together editing use strict CSR,
    whereas Image editing uses raster-tolerant success.
    }
    \label{fig:overall_performance}
\end{figure}

Figure~\ref{fig:overall_performance} summarizes overall QA and editing
performance across models and input modalities. QA accuracy is averaged over
the open-ended and multiple-choice settings. Under GeoJSON, GPT-6 Astra
achieves the strongest overall performance, with 96.5\% QA accuracy and
97.6\% editing success. KIMI K3 follows at 90.7\% and 85.1\%, respectively.
Performance then declines across the remaining models, although the relative
order is not identical between QA and editing.

QA and editing performance are broadly aligned under GeoJSON. Models with
higher QA accuracy generally also achieve higher editing success, but several
exceptions remain. MiMo-V2.6-Pro reaches 67.3\% QA accuracy, above
Qwen3.8-27B at 55.5\%, whereas Qwen3.8-27B performs better on editing
(54.4\% vs.\ 43.0\%). Thus, strong relation understanding does not always
translate into equally strong constrained editing.

Performance also differs across input modalities. Map image scores are
generally lower than GeoJSON scores for both QA and spatial editing. For
MiMo-V2.6-Pro, mean QA accuracy decreases from 67.3\% with GeoJSON to
45.2\% with map image, while Together reaches 65.2\%. In spatial editing,
the same model achieves 43.0\% CSR with GeoJSON and 51.4\% with Together,
whereas its raster-tolerant success rate with map image is 16.5\%.
Together therefore remains close to GeoJSON in QA, but its effect on spatial
editing varies across models. Because map image editing uses a different
output interface and evaluation protocol, its score is not directly comparable
to strict CSR.

\subsection{Performance across Relations and Cities}

\begin{figure}[t]
    \centering
    \includegraphics[width=\linewidth]{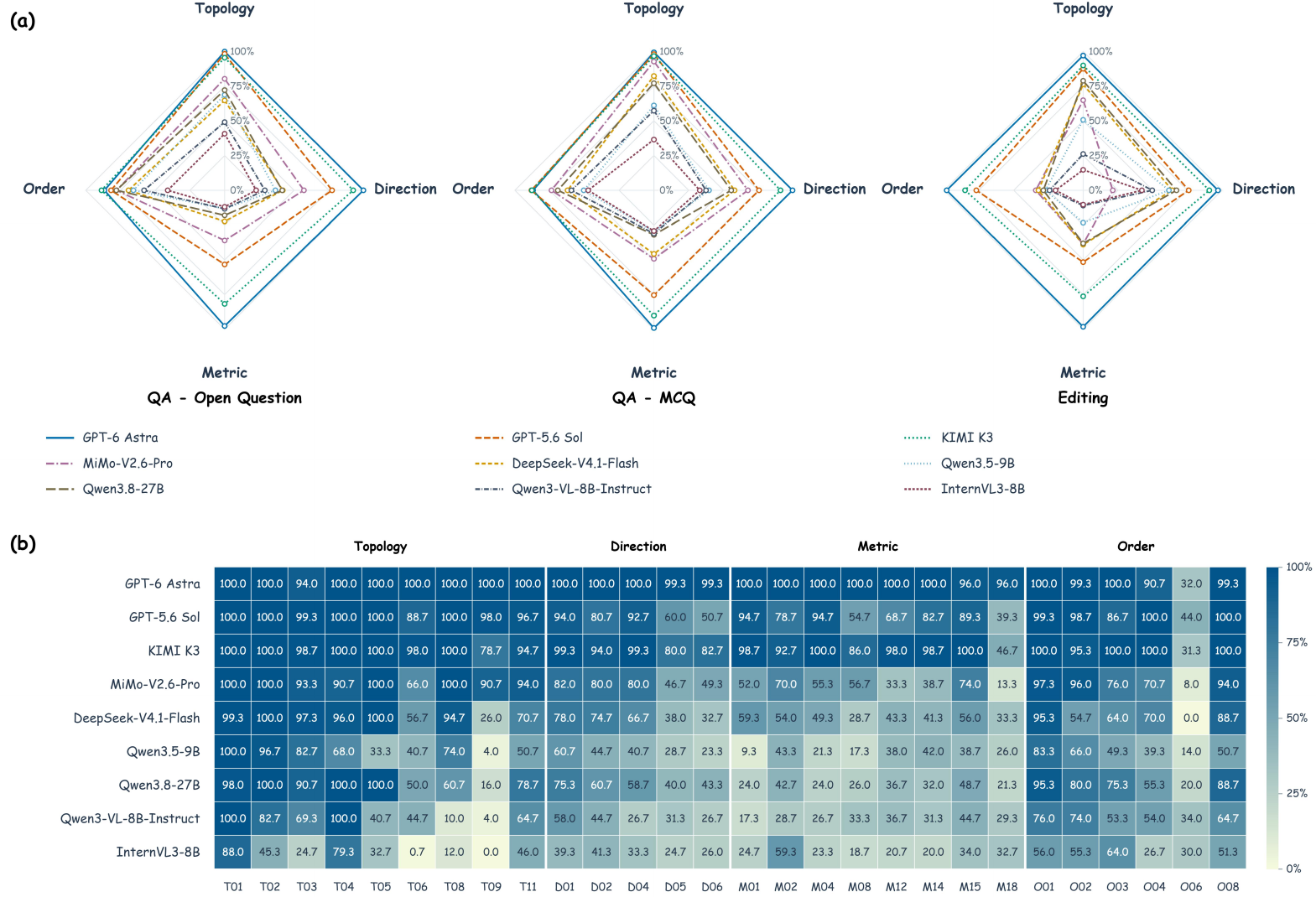}
    \caption{
    Performance across spatial relations under GeoJSON input.
    (a) Mean performance across the four relation families for open-ended QA,
    multiple-choice QA, and spatial editing.
    (b) Relation-level multiple-choice QA accuracy for each model across the
    28 spatial relations. Metric relations are generally more challenging,
    while QA and spatial editing show different patterns across relation families.
    }
    \label{fig:relation_heatmap}
\end{figure}

As shown in Figure~\ref{fig:relation_heatmap}(a), performance varies
substantially across the four relation families under GeoJSON. In open-ended
QA, topology and ordering achieve 74.4\% and 71.9\% accuracy, followed by
direction at 55.3\% and metric at 38.6\%. Multiple-choice QA shows a similar
pattern, with 77.9\% for topology, 69.4\% for ordering, 62.0\% for direction,
and 53.4\% for metric. Spatial editing shows a different distribution.
Topology and direction achieve the highest CSR at 65.1\% and 63.4\%,
whereas ordering falls to 47.3\% and metric remains the lowest at 42.9\%.

Figure~\ref{fig:relation_heatmap}(b) further shows substantial variation
among individual relations in multiple-choice QA. Relation codes used in the
figure are listed in Appendix~B, Table~4. Averaged across models, Equality
(T01) achieves 98.4\% accuracy and Touching (T04) reaches 92.7\%, whereas
Boundary Coverage (T09) drops to 46.4\%. Metric relations are generally more
difficult: POI Density (M18) achieves 37.6\%, and Hausdorff Distance (M08) reaches 46.8\%. Ordering relations also vary widely, from 89.2\% for Along-Line Order
(O01) to 23.7\% for Layer Hierarchy (O06). Layer Hierarchy is
attribute-dependent, and the required road-layer attribute is omitted from the
geometry-only GeoJSON QA input. Its raw accuracy therefore reflects information
availability in addition to geometric reasoning difficulty.

Performance varies relatively little across the six evaluated cities.
For open-ended GeoJSON QA, the within-model range across cities is
1.00--5.71 percentage points. By comparison, pooled accuracy differs by
67.03 percentage points between the highest- and lowest-performing models.
The observed variation across models is therefore substantially larger than
the within-model variation across these six cities. These results do not
establish generalization to unseen cities.

\subsection{Error Analysis and Discussion}
\label{sec:error-analysis}

\begin{figure}[t]
    \centering
    \includegraphics[width=\linewidth]{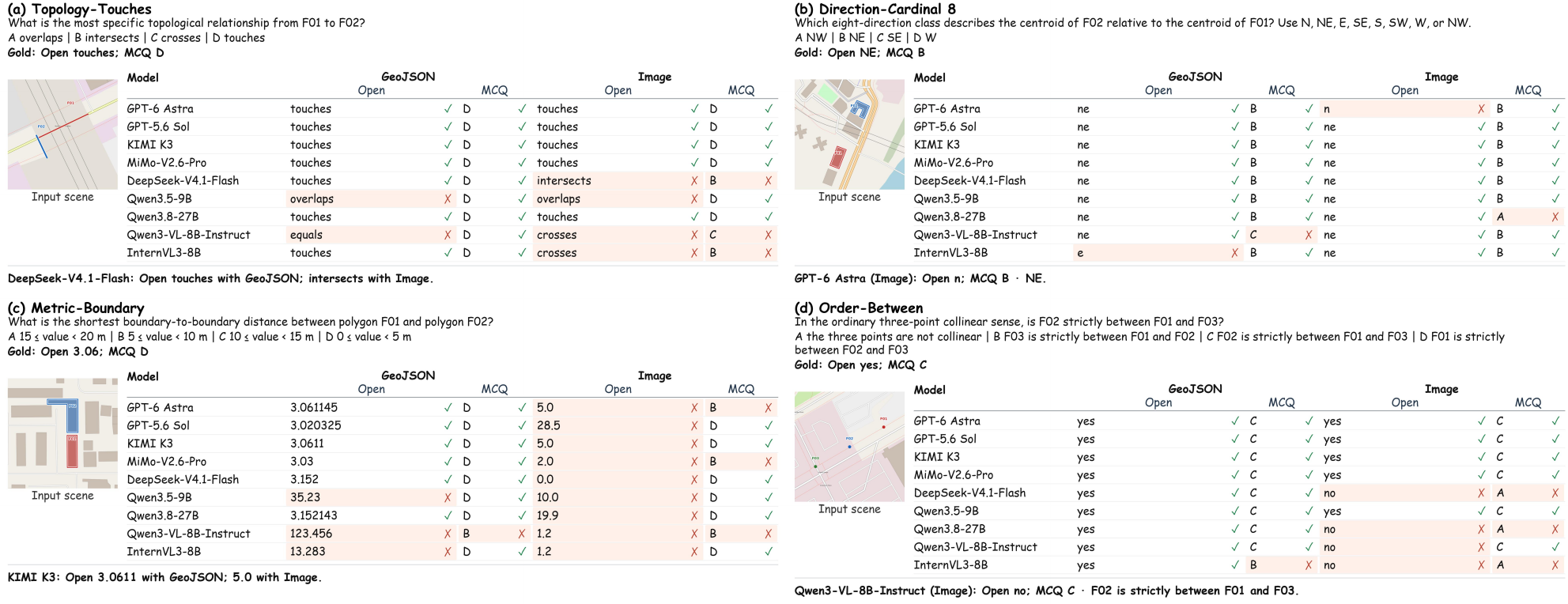}
    \caption{
    Representative QA cases across the four spatial relation families.
    Map images often preserve coarse spatial judgments but are less reliable
    for precise boundary and metric reasoning.
    }
    \label{fig:qualitative_cases}
\end{figure}

\begin{figure}[t]
    \centering
    \includegraphics[width=\linewidth]{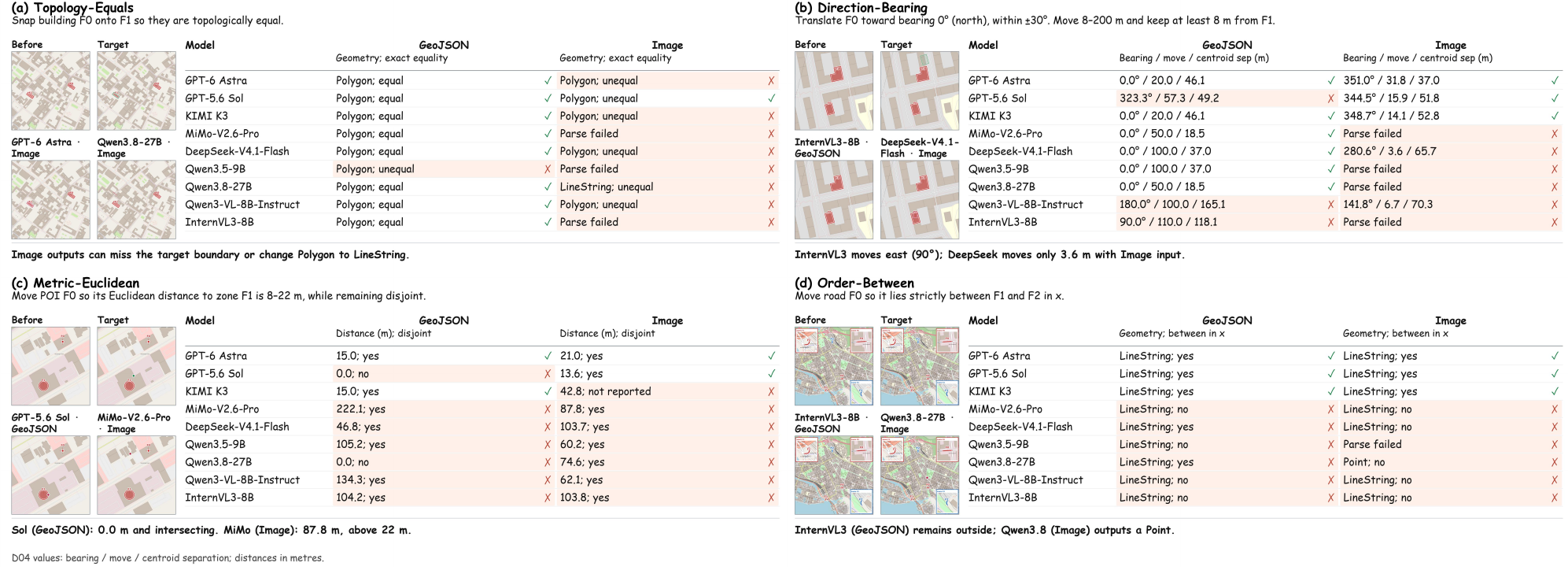}
    \caption{
    Representative spatial editing cases across four relation families,
    illustrating failures in geometric execution and constraint satisfaction.
    }
    \label{fig:editing_cases}
\end{figure}

We further examine whether spatial-relation recognition transfers to spatial
intervention, which relations remain difficult, and how representation affects
spatial reasoning. In GeoSpatialBench, this
capability is operationalized through spatial editing. Figures~\ref{fig:qualitative_cases} and
\ref{fig:editing_cases} show representative QA and editing cases, while
complete relation-level statistics are provided in
Appendix~\ref{app:error-analysis}.

The representative cases reveal distinct sources of error across tasks and
representations. In QA, some relations remain stable across inputs, while
precise metric estimates are more fragile. In the Touching case,
DeepSeek-V4.1-Flash is correct with GeoJSON but predicts \emph{intersects}
from the map image. For Boundary Distance, KIMI K3 gives an accurate
open-ended value with GeoJSON but fails under the map image, while still
selecting the correct multiple-choice interval. Spatial editing introduces a different class of errors. In the Bearing case, KIMI K3 succeeds under both inputs, whereas InternVL3-8B moves the target in the wrong direction with GeoJSON. In the Euclidean Distance case, GPT-6 Astra succeeds under both inputs, while KIMI K3 succeeds with GeoJSON but produces a 42.8\,m distance from the map image, outside the required 8--22\,m range.

\paragraph{Relation recognition can approach saturation before spatial intervention.}
Several basic topological relations are already recognized reliably in
GeoJSON MCQ-4: Equality, Touching, and Disjointness have error rates of only
1.6\%, 7.3\%, and 8.4\%. However, this does not consistently transfer to
editing. Touching and Disjointness increase to 71.2\% and 47.3\% editing
error, whereas Equality remains easy at 3.3\%. Thus, recognizing a relation
and constructing a geometry that satisfies it are distinct capabilities.
Figure~\ref{fig:editing_cases} also illustrates this gap: models may choose a
plausible operation but still produce an incorrect direction, distance, or
final geometry. Editing therefore remains discriminative even when some
recognition tasks approach saturation. The gap is also relation-dependent. For direction relations, GeoJSON editing reaches 63.4\% CSR, close to the 62.0\% MCQ-4 accuracy.

\paragraph{Metric reasoning remains the most persistent bottleneck.}
Metric relations remain the most difficult family under both GeoJSON and
map-image inputs. Even with explicit GeoJSON geometry, open-ended metric error
reaches 61.4\%. Several numerical relations are particularly difficult:
Minimum Distance, Hausdorff Distance, Total Perimeter, and POI Density exceed
70\% error in open-ended QA. Figure~\ref{fig:qualitative_cases} shows a related
pattern. Models can sometimes identify the correct distance interval while
failing to produce an accurate numerical value. These results point to a
persistent weakness in quantitative geometric reasoning. Access to explicit
coordinates reduces representational ambiguity, but does not by itself ensure
reliable computation of continuous spatial quantities.

\paragraph{Explicit geometry better supports precise spatial reasoning.}
For QA, GeoJSON and map-image conditions use the same scene, question, and
reference answer, enabling a direct comparison of representations. GeoJSON
has lower family-level error across all four relation families, with the
largest gap in metric reasoning. Figure~\ref{fig:qualitative_cases} shows the
same tendency: coarse directional or categorical judgments may remain correct
from maps, while precise boundary and distance estimates are less reliable.
Together inputs also do not consistently improve over GeoJSON alone,
suggesting that additional visual context does not automatically strengthen
an explicit geometric representation. For editing, the GeoJSON--image gap
should be interpreted cautiously because image editing also uses a different
output interface.

\paragraph{Toward explicit spatial representations.}
These patterns suggest that precise spatial reasoning may require
representations that preserve geometric and relational structure more
explicitly. Metric reasoning depends on continuous quantities, while
rasterization requires exact geometry to be recovered indirectly from a
finite-resolution image. Our experiments do not isolate a specific
architectural cause, but they suggest that additional multimodal visual
information alone may be insufficient. Spatial tokenization methods such as
SPOK \citep{huang2025spok} provide one possible direction by representing
geographic space and spatial relations explicitly. Such representations may
help bridge spatial recognition, geometric reasoning, and spatial
intervention.

\section{Conclusion}
\label{sec:conclusion}

We introduced GeoSpatialBench, a benchmark for evaluating spatial relation understanding, geometric computation, and constrained spatial editing in urban geographic scenes using structured GeoJSON and rendered map images. Across nine models, GeoJSON generally yields stronger performance than map images, while combining the two inputs does not consistently improve performance. Precise metric computation and boundary relations remain challenging, and constraint violations account for most editing failures. By evaluating spatial reasoning and editing within a common, automatically verifiable framework, GeoSpatialBench shows that recognizing a spatial relation does not necessarily mean a model can construct geometry that satisfies it. A promising next step is to test whether dedicated, relation-aware representations can help bridge this gap. Difficulty with basic 2D geometric constraints also highlights the challenge of more complex spatial generation and provides a baseline for future work.

\section*{Reproducibility Statement}
Section~\ref{sec:methods} specifies the six OpenStreetMap source cities, instance counts, GeoJSON and map-image construction, task formats, and GIS-based procedures for generating answers and checking edited geometry. It also defines the parsing rules, numerical tolerances, corrective retry, and separate raster-tolerant criterion for image editing. The evaluation code and supporting benchmark materials will be provided by the authors upon request. 

\section*{AI Use Statement}
AI-based tools were used to assist with language editing and \LaTeX{} formatting. The authors used OpenAI's ChatGPT (5.6 Sol) to assist with code development, data processing, and language editing of the manuscript. The authors reviewed and tested the AI-assisted code, checked the data-processing outputs, and reviewed and revised the AI-edited text. The authors take full responsibility for the accuracy and integrity of the methods, results, and final manuscript.

\bibliography{references}
\bibliographystyle{geospatialbench}

\clearpage
\appendix
\section{Additional Results}
\label{app:additional-results}
Table~\ref{tab:main_results} jointly reports GeoJSON, Image, and Together
performance for all nine models, ordered by six-city GeoJSON open-ended QA.
Table~\ref{tab:city-performance} shows the six-city
variation in GeoJSON open-ended QA. The differences between cities range from 1.00\% to 5.71\%. 

\begin{table}[H]
\centering
\caption{Overall performance (\%; higher is better), ordered by GeoJSON open-ended QA. Each model uses 4,200 instances across six cities. Image editing reports raster-tolerant success; GeoJSON and Together editing report strict CSR. Bold marks the highest reported score in each column.}
\label{tab:main_results}
\scriptsize
\setlength{\tabcolsep}{2.2pt}
\resizebox{\textwidth}{!}{%
\begin{tabular}{@{}lrrrrrrrrr@{}}
\toprule
 & \multicolumn{3}{c}{GeoJSON} & \multicolumn{3}{c}{Map image} & \multicolumn{3}{c}{Together} \\
\cmidrule(lr){2-4}\cmidrule(lr){5-7}\cmidrule(l){8-10}
Model & Open & MCQ-4 & Edit & Open & MCQ-4 & Edit & Open & MCQ-4 & Edit \\
\midrule
GPT-6 Astra & \textbf{96.36} & \textbf{96.64} & \textbf{97.60} & \textbf{60.57} & \textbf{73.12} & \textbf{44.81} & \textbf{94.86} & \textbf{95.86} & \textbf{98.71} \\
KIMI K3 & 89.52 & 91.88 & 85.10 & 51.43 & 59.29 & 39.29 & 92.86 & 93.43 & 85.29 \\
GPT-5.6 Sol & 78.12 & 85.43 & 72.90 & 47.40 & 56.02 & 30.12 & 77.14 & 84.29 & 70.86 \\
MiMo-V2.6-Pro & 62.98 & 71.71 & 42.98 & 39.19 & 51.14 & 16.50 & 59.55 & 70.93 & 51.38 \\
Qwen3.8-27B & 52.45 & 58.64 & 54.38 & 38.43 & 42.00 & 18.29 & 53.52 & 55.93 & 52.19 \\
DeepSeek-V4.1-Flash & 49.50 & 63.17 & 54.02 & 33.69 & 43.10 & 16.74 & 52.79 & 59.48 & 52.60 \\
Qwen3.5-9B & 46.50 & 45.95 & 39.79 & 35.52 & 43.55 & 6.14 & 42.62 & 47.21 & 29.71 \\
Qwen3-VL-8B-Instruct & 37.19 & 46.69 & 25.62 & 34.86 & 40.90 & 9.55 & 42.48 & 47.57 & 21.55 \\
InternVL3-8B & 29.33 & 36.07 & 19.50 & 27.55 & 34.14 & 11.67 & 28.98 & 34.36 & 9.93 \\
\bottomrule
\end{tabular}%
}
\end{table}

\begin{table}[H]
\centering
\caption{Open-ended GeoJSON QA across six cities (\%; higher is better), ordered by pooled accuracy. Each city has 700 questions per model. Overall pools 4,200 questions; Range is the maximum minus minimum city score in percentage points.}
\label{tab:city-performance}
\small
\setlength{\tabcolsep}{2.5pt}
\begin{tabular*}{\linewidth}{@{\extracolsep{\fill}}lrrrrrrrr@{}}
\toprule
Model & Stockholm & Paris & Nairobi & Hong Kong & Wuhan & Detroit & Overall & Range\\
\midrule
GPT-6 Astra & 95.57 & 96.14 & 96.14 & 96.71 & 96.86 & 96.71 & 96.36 & 1.29 \\
KIMI K3 & 89.29 & 90.00 & 89.86 & 89.43 & 89.00 & 89.57 & 89.52 & 1.00 \\
GPT-5.6 Sol & 77.86 & 79.57 & 79.57 & 77.57 & 76.57 & 77.57 & 78.12 & 3.00 \\
MiMo-V2.6-Pro & 62.57 & 64.14 & 62.14 & 64.43 & 61.43 & 63.14 & 62.98 & 3.00 \\
Qwen3.8-27B & 51.14 & 53.14 & 53.00 & 52.43 & 51.71 & 53.29 & 52.45 & 2.14 \\
DeepSeek-V4.1-Flash & 48.43 & 49.00 & 49.14 & 48.71 & 49.86 & 51.86 & 49.50 & 3.43 \\
Qwen3.5-9B & 44.14 & 48.00 & 46.57 & 47.00 & 46.86 & 46.43 & 46.50 & 3.86 \\
Qwen3-VL-8B-Instruct & 37.00 & 36.71 & 38.43 & 37.00 & 38.43 & 35.57 & 37.19 & 2.86 \\
InternVL3-8B & 26.14 & 31.86 & 31.00 & 29.00 & 30.00 & 28.00 & 29.33 & 5.71 \\
\bottomrule
\end{tabular*}
\end{table}

\clearpage
\section{Additional Error Analysis}
\label{app:error-analysis}

\subsection{Distribution by Relation}
\label{app:error-overall}

\begin{figure}[H]
    \centering
    \includegraphics[width=\textwidth,height=0.76\textheight,keepaspectratio]{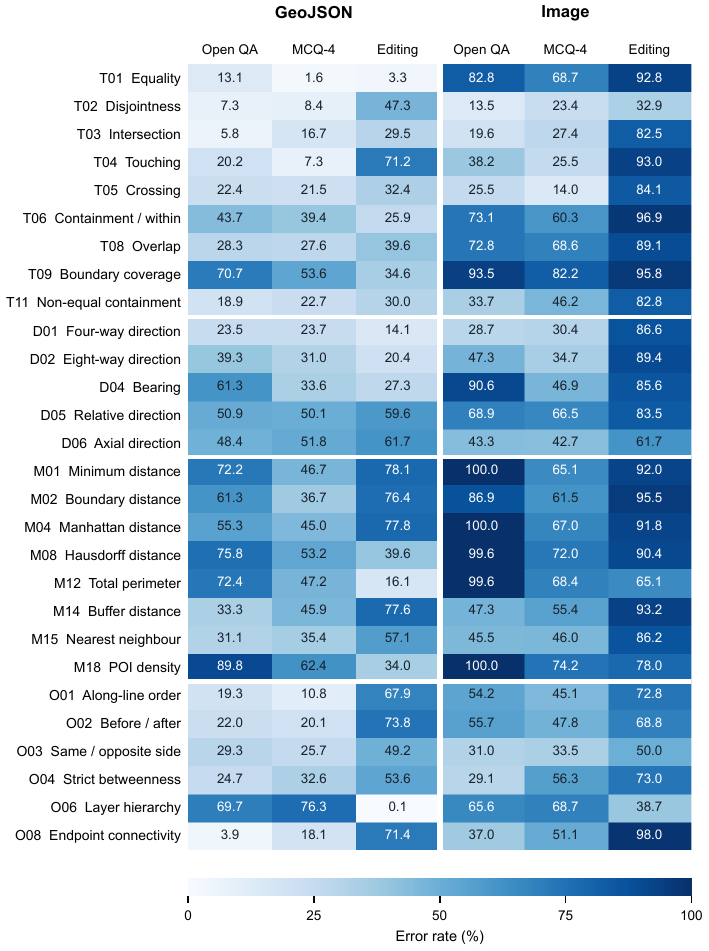}
    \caption{Error rates by spatial relation, input, and task (\%; lower is better).}
    \label{fig:error-overall-distribution}
\end{figure}

Figure~\ref{fig:error-overall-distribution} and
Tables~\ref{tab:error-six-city-family} and~\ref{tab:error-six-city-relations}
summarize relation-level error rates for the nine reported models.

\clearpage
\begin{table}[H]
\centering
\caption{Error rates by relation family (\%; lower is better), pooled over the nine reported models. GeoJSON and Map Image use six cities for every model.}
\label{tab:error-six-city-family}
\small
\setlength{\tabcolsep}{5pt}
\begin{tabular}{lrrrrrr}
\toprule
& \multicolumn{3}{c}{GeoJSON} & \multicolumn{3}{c}{Map image} \\
\cmidrule(lr){2-4}\cmidrule(l){5-7}
Family & Open QA & MCQ-4 & Editing & Open QA & MCQ-4 & Editing \\
\midrule
Topology & 25.61 & 22.09 & 34.86 & 50.30 & 46.27 & 83.32 \\
Direction & 44.68 & 38.04 & 36.64 & 55.76 & 44.21 & 81.37 \\
Metric & 61.40 & 46.57 & 57.09 & 84.87 & 63.69 & 86.53 \\
Ordering & 28.14 & 30.59 & 52.67 & 45.43 & 50.44 & 66.87 \\
\bottomrule
\end{tabular}
\end{table}

\begin{table}[H]
\centering
\caption{Error rates for the 28 spatial relations (\%; lower is better).}
\label{tab:error-six-city-relations}
\scriptsize
\setlength{\tabcolsep}{3.5pt}
\resizebox{\textwidth}{!}{%
\begin{tabular}{lllrrrrrr}
\toprule
&&& \multicolumn{3}{c}{GeoJSON} & \multicolumn{3}{c}{Map image} \\
\cmidrule(lr){4-6}\cmidrule(l){7-9}
ID & Relation & Family & Open QA & MCQ-4 & Editing & Open QA & MCQ-4 & Editing \\
\midrule
T01 & Equality & Topology & 13.11 & 1.63 & 3.26 & 82.78 & 68.73 & 92.82 \\
T02 & Disjointness & Topology & 7.33 & 8.37 & 47.26 & 13.47 & 23.43 & 32.90 \\
T03 & Intersection & Topology & 5.78 & 16.67 & 29.48 & 19.59 & 27.43 & 82.53 \\
T04 & Touching & Topology & 20.22 & 7.33 & 71.19 & 38.20 & 25.47 & 92.98 \\
T05 & Crossing & Topology & 22.44 & 21.48 & 32.44 & 25.47 & 14.04 & 84.08 \\
T06 & Containment / within & Topology & 43.70 & 39.41 & 25.93 & 73.14 & 60.33 & 96.90 \\
T08 & Overlap & Topology & 28.30 & 27.63 & 39.63 & 72.82 & 68.57 & 89.06 \\
T09 & Boundary coverage & Topology & 70.74 & 53.63 & 34.59 & 93.55 & 82.20 & 95.84 \\
T11 & Non-equal containment & Topology & 18.89 & 22.67 & 30.00 & 33.71 & 46.20 & 82.78 \\
\midrule
D01 & Four-way direction & Direction & 23.48 & 23.70 & 14.07 & 28.73 & 30.37 & 86.61 \\
D02 & Eight-way direction & Direction & 39.26 & 31.04 & 20.44 & 47.27 & 34.69 & 89.39 \\
D04 & Bearing & Direction & 61.33 & 33.56 & 27.33 & 90.61 & 46.86 & 85.63 \\
D05 & Relative direction & Direction & 50.89 & 50.15 & 59.63 & 68.90 & 66.45 & 83.51 \\
D06 & Axial direction & Direction & 48.44 & 51.78 & 61.70 & 43.27 & 42.69 & 61.71 \\
\midrule
M01 & Minimum distance & Metric & 72.15 & 46.67 & 78.07 & 100.00 & 65.06 & 92.00 \\
M02 & Boundary distance & Metric & 61.33 & 36.74 & 76.37 & 86.94 & 61.47 & 95.51 \\
M04 & Manhattan distance & Metric & 55.26 & 45.04 & 77.85 & 100.00 & 67.02 & 91.84 \\
M08 & Hausdorff distance & Metric & 75.78 & 53.19 & 39.63 & 99.59 & 72.00 & 90.37 \\
M12 & Total perimeter & Metric & 72.44 & 47.19 & 16.07 & 99.59 & 68.41 & 65.06 \\
M14 & Buffer distance & Metric & 33.26 & 45.93 & 77.63 & 47.27 & 55.43 & 93.22 \\
M15 & Nearest neighbour & Metric & 31.11 & 35.41 & 57.11 & 45.55 & 45.96 & 86.20 \\
M18 & POI density & Metric & 89.85 & 62.44 & 34.00 & 100.00 & 74.20 & 78.04 \\
\midrule
O01 & Along-line order & Ordering & 19.33 & 10.81 & 67.93 & 54.20 & 45.06 & 72.82 \\
O02 & Before / after & Ordering & 22.00 & 20.07 & 73.85 & 55.67 & 47.84 & 68.82 \\
O03 & Same / opposite side & Ordering & 29.26 & 25.70 & 49.19 & 31.02 & 33.55 & 49.96 \\
O04 & Strict betweenness & Ordering & 24.67 & 32.59 & 53.56 & 29.06 & 56.33 & 72.98 \\
O06 & Layer hierarchy & Ordering & 69.70 & 76.30 & 0.07 & 65.63 & 68.73 & 38.69 \\
O08 & Endpoint connectivity & Ordering & 3.85 & 18.07 & 71.41 & 36.98 & 51.10 & 97.96 \\
\bottomrule
\end{tabular}%
}
\end{table}

\end{document}